\pdfoutput=1

\documentclass[11pt]{article}

\usepackage{acl}

\usepackage{times}
\usepackage{latexsym}
\usepackage{adjustbox} 

\usepackage[T1]{fontenc}

\usepackage[utf8]{inputenc}

\usepackage{microtype}

\usepackage{inconsolata}

\title{MasonTigers at SemEval-2024 Task 10: Emotion Discovery and Flip Reasoning in Conversation with Ensemble of Transformers and Prompting}

 \author{Al Nahian Bin Emran, Amrita Ganguly, Sadiya Sayara Chowdhury Puspo, \\ {\bf Nishat Raihan, Dhiman Goswami} \\ George Mason University, USA \\
 \texttt{abinemra@gmu.edu}
 }
 
\begin{document}
\maketitle
\begin{abstract}
In this paper, we present \textit{MasonTigers}' participation in SemEval-2024 Task 10, a shared task aimed at identifying emotions and understanding the rationale behind their flips within monolingual English and Hindi-English code-mixed dialogues. This task comprises three distinct subtasks – emotion recognition in conversation for Hindi-English code-mixed dialogues, emotion flip reasoning for Hindi-English code-mixed dialogues, and emotion flip reasoning for English dialogues. Our team, \textit{MasonTigers}, contributed to each subtask, focusing on developing methods for accurate emotion recognition and reasoning. By leveraging our approaches, we attained impressive F1-scores of 0.78 for the first task and 0.79 for both the second and third tasks. This performance not only underscores the effectiveness of our methods across different aspects of the task but also secured us the top rank in the first and third subtasks, and the $2^{nd}$ rank in the second subtask. Through extensive experimentation and analysis, we provide insights into our system's performance and contributions to each subtask.

\end{abstract}

\section{Introduction}
Emotion Recognition in Conversation (ERC) has emerged as a crucial area of research within Natural Language Processing (NLP). Its primary objective is to understand and replicate human emotions, thereby placing it at the forefront of NLP research endeavors. The focus on ERC arises from the crucial role of understanding human emotions in the broader field of artificial intelligence. The notable contributions of researchers like \cite{Ekman, Picard97, hazarika-etal-2018a, 
 hazarika-etal-2018-conversational-b, zhong2019knowledgeenriched, ghosal2019dialoguegcn, jiao2019higru} have played a pivotal role in advancing this field. Their work has contributed to shaping the landscape of ERC research, underscoring its importance and providing valuable insights into the intricate mechanisms of human emotion recognition within conversational contexts. 
Recognizing emotions in conversations is also crucial, especially when emotions suddenly change. However, just spotting these changes is not enough; we need to understand what causes them, like in customer service situations. This understanding helps make dialogue systems better at handling emotions, which improves the experience for users. Emotion-Flip Reasoning (EFR), as described by \citet{kumar2021discovering}, is a groundbreaking effort to figure out what triggers emotional flips in group conversations. This task is not just about spotting emotional changes but also understand the words that lead up to those changes.

This shared task is designed to advance research in Emotion Recognition in Conversation (ERC) and Emotion-Flip Reasoning (EFR). It consists of three distinct subtasks aimed at exploring these areas comprehensively.

The goal of the first subtask is to perform ERC in Hindi-English code-mixed conversations. Participants are tasked with tagging each utterance in a multiparty code-mixed conversation with one of the eight emotion labels: anger, disgust, fear, sadness, surprise, joy, contempt, and neutral. This task is challenging due to the nature of code-mixed language, which combines linguistic elements from both Hindi and English, leading to unique syntactic and semantic structures. The ability to accurately recognize emotions in such mixed-language settings is critical for applications in social media monitoring, customer service, and human-computer interaction, where code-mixing is prevalent. The dataset for this task includes dialogues from various contexts, reflecting the spontaneous and informal use of language, which adds another layer of complexity to emotion recognition.

The second subtask focuses on identifying trigger utterances for emotion flips in Hindi-English code-mixed conversations, involving understanding speaker interactions and context. This task is vital for applications like customer service and user experience enhancement. The dataset reflects cultural and linguistic dynamics. The third subtask is similar but in monolingual English, emphasizing the identification of triggers in English dialogues. Our approach secured first place in Subtasks A and C, and second place in Subtask B, with F1-scores of 0.83, 0.81, and 0.86 on evaluation sets, respectively, demonstrating our methodology's effectiveness.

\section{Related Work}
Research into identifying emotions has been an ongoing endeavor, initially focusing on understanding emotions in standalone contexts. There are multiple research that paves the way in this domain by exploring basic emotion theories and effective computing. \citet{Ekman}'s foundational work on facial expressions and basic emotions laid the groundwork for emotion recognition systems, while \citet{Picard97}'s introduction of affective computing highlighted the importance of emotional interaction between humans and machines. Moreover, \citet{cui2020eeg} extended these concepts by integrating EEG signals into emotion recognition, demonstrating the potential of multimodal approaches.

Recognizing the importance of contextual cues, the spotlight shifted towards emotion detection within conversations, particularly in Emotion Recognition in Conversation (ERC). Early attempts at ERC involved heuristic methods and traditional machine learning techniques. \cite{li2007research} focused on rule-based systems to identify emotions from textual cues, while \citet{fitrianie2003multi} used pattern recognition techniques for emotion classification in spoken dialogues. However, these methods had limitations in capturing the complexities of human emotions in conversations.

The advent of deep learning has revolutionized ERC, with various models leveraging neural networks to enhance emotion detection. \citet{zhong2019knowledgeenriched} introduced knowledge-enriched attention networks for ERC, incorporating external knowledge to improve emotion classification. \citet{hazarika-etal-2018a} proposed conversational memory networks, which utilize hierarchical attention mechanisms to model speaker-specific information. Recent studies have explored transformer-based architectures and graph neural networks to capture contextual dependencies and speaker interactions in dialogues \cite{li2022bieru,yang2022hybrid,tu2022context,ma2022multi}.

While recent studies have examined emotion analysis in code-mixed language, the focus remains primarily on social media texts \cite{ilyas2023emotion,wadhawan2021towards,sasidhar2020emotion} and reviews \cite{zhu2022leveraging,suciati2020aspect}. These works have highlighted the unique challenges of code-mixed text, such as language switching and informal expressions. Despite some exploration into aspects such as sarcasm \cite{kumar2022did, kumar2022explaining}, offense \cite{madhu2023detecting}, and humor \cite{bedi2021multi} within code-mixed conversations, the field of emotion analysis remains largely unexplored, lacking sufficient literature. This shared task aims to bridge this gap by delving into the under-explored domain of ERC, specifically within Hindi-English code-mixed dialogues. Additionally, recent developments have introduced valuable code-mixed datasets such as OffMix-3L \cite{goswami2023offmix}, SentMix-3L \cite{raihan2023sentmix}, EmoMix-3L \cite{raihan2024emomix} and TB-OLID \cite{raihan2023offensive}, facilitating advancements in this area.

The exploration of emotions within linguistic contexts remains an area of research with limited exploration. Few studies have ventured into this domain, aiming to unravel the underlying causes of expressed emotions in text, often referred to as Emotion-Cause Analysis. \citet{lee2010text} focused on identifying textual cues that trigger specific emotions, while \citet{wang2022multimodal} integrated multimodal data to enhance emotion-cause analysis. While the concept of emotion-cause analysis and emotion-flip reasoning tasks may appear interconnected in theory, they diverge significantly in practice. Emotion-cause analysis focuses on identifying phrases within text that trigger specific emotions, whereas the Emotion Flip Reasoning (EFR) task involves extracting causes or triggers behind emotional flips in conversational dialogues involving multiple speakers \cite{kumar2023emotion}. These triggers comprise one or more utterances from the dialogue history, highlighting the dynamic nature of emotions in multi-party interactions.

\section{Datasets}

\begin{table*}[!ht]
\centering
\small
\begin{tabular}{lcccccccc|c|c}
\hline
& Disgust & Joy & Surprise & Anger & Fear & Neutral & Sadness & Contempt & \# of Dialogue & \# of Utterance\\ 
\hline
Train   &  127       &  1596   &  441        &   819    &  514    & 3909        & 558        &    542      &  8506    & 343\\ 
Dev    &  21       &  228   &   66       & 118  & 88   &    633  &  126       & 74        & 1354         &      46 \\ 
Test   &  17       &  349   &   57       &   142    &  122    &   656      &   155      &     82     &  1580   & 57 \\ 
\hline
\end{tabular}
\caption{Dataset Subtask A}
\label{tab:emotion_analysisA}
\end{table*}

\begin{table*}[!ht]
\centering
\begin{tabular}{lcccc}
\hline
& Dialogue w/o trigger (0) & Dialogue with Trigger (1) & \# of Dialogue & \# of Utterance\\ 
\hline
Train   &    92235     & 6542    & 98777 & 4893 \\ 
Dev    &    7028     &  434   & 7462  & 389  \\ 
Test   &   7274      & 416    & 7690  &  385 \\ 
\hline
\end{tabular}
\caption{Dataset Subtask B}
\label{tab:emotion_analysisB}
\end{table*}

\begin{table*}[!ht]
\centering
\begin{tabular}{lcccc}
\hline
& Dialogue w/o trigger (0) & Dialogue with Trigger (1) & \# of Dialogue & \# of Utterance\\ 
\hline
Train   &   29423      &  5577   & 35000 & 4000 \\ 
Dev    &   3028      &  494   & 3522  &  389 \\ 
Test   &     7473    &   1169  &  8642 &  1002 \\ 
\hline
\end{tabular}
\caption{Dataset Subtask C}
\label{tab:emotion_analysisC}
\end{table*}

The EDiReF shared task at SemEval 2024 comprises three subtasks: (i) Emotion Recognition in Conversation (ERC) in Hindi-English code-mixed conversations, (ii) Emotion Flip Reasoning (EFR) in Hindi-English code-mixed conversations, and (iii) EFR in English conversations.

The dataset for subtask 1, Emotion Recognition in Conversation (ERC) for code-mixed dialogues, consists of 11,440 dialogues derived from 446 utterances. It is segmented into training, development (dev), and test sets. Specifically, the training set comprises 8,506 dialogues from 343 utterances, the development set includes 1,354 dialogues from 46 utterances, and the test set contains 1,580 dialogues from 57 utterances. Each dialogue is annotated with one of eight emotions: anger, disgust, fear, sadness, surprise, joy, contempt, and neutral. The specifics of the dataset for this track can be found in Table \ref{tab:emotion_analysisA}.

Subtask 2, Emotion Flip Reasoning (EFR) in Hindi-English code-mixed conversations, focuses on identifying emotion triggers at the dialogue level. The dataset includes a total of 113,929 dialogues and 5,667 utterances across training, development, and test sets. In the training set, there are 98,777 dialogues, with 6,542 dialogues containing emotion triggers (labeled as 1), and 92,235 dialogues without triggers (labeled as 0), encompassing 4,893 utterances. The development set includes 7,462 dialogues, of which 434 dialogues contain triggers and 7,028 do not, with 389 utterances annotated. Similarly, the test set consists of 7,690 dialogues, with 416 dialogues containing triggers and 7,274 dialogues without, totaling 385 utterances. The details of the dataset of this track is available in Table \ref{tab:emotion_analysisB}.

Subtask 3, Emotion Flip Reasoning (EFR) in English conversations, aims to pinpoint specific utterances triggering emotional shifts in multi-party dialogues. The dataset comprises a total of 47,164 dialogues and 5,391 utterances across training, development, and test sets. The training set includes 35,000 dialogues, with 5,577 dialogues containing triggers (labeled as 1) and 29,423 dialogues without triggers (labeled as 0), covering 4,000 utterances. In the development set, there are 3,522 dialogues, with 494 containing triggers and 3,028 without, annotated with 389 utterances. The test set consists of 8,642 dialogues, with 1,169 containing triggers and 7,473 without, encompassing 1002 utterances. This dataset supports the investigation of emotional triggers in English conversations, contributing to advancements in emotion reasoning within dialogue analysis. The details of the dataset of this track is available in Table \ref{tab:emotion_analysisC}.

\section{Experiments}

In this section, we describe the experimental setup and the results obtained for the three subtasks: Emotion Recognition in Conversation (ERC) in Hindi-English code-mixed dialogues, Emotion Flip Reasoning (EFR) in Hindi-English code-mixed dialogues, and EFR in English dialogues.

The dataset for Subtask A consists of dialogues annotated with eight emotion categories: disgust, joy, surprise, anger, fear, neutral, sadness, and contempt. The training set contains 8,506 dialogues from 343 utterances, the development set includes 1,354 dialogues from 46 utterances, and the test set comprises 1,580 dialogues from 57 utterances (Table \ref{tab:emotion_analysisA}). We experimented with several models, including MuRIL \cite{khanuja2021muril}, XLM-R \cite{conneau2019unsupervised}, mBERT \cite{devlin2018bert}, HingBERT \cite{jain2021hingbert}, and IndicBERT \cite{kakwani2020indicnlpsuite}. Our approach utilized a weighted ensemble of these models to improve performance. The results, as shown in Table \ref{table:sub-task1_results}, indicate that the weighted ensemble achieved the highest F1-scores on both the evaluation (0.83) and test (0.78) sets, outperforming individual models.

For Subtask B, the dataset consists of dialogues labeled with the presence or absence of triggers for emotion flips. The training set includes 98,777 dialogues, the development set comprises 7,462 dialogues, and the test set contains 7,690 dialogues (Table \ref{tab:emotion_analysisB}). We evaluated multiple models, including MuRIL \cite{khanuja2021muril}, XLM-R \cite{conneau2019unsupervised}, mBERT \cite{devlin2018bert}, FLAN-T5 \cite{raffel2020exploring}, and GPT-4 Turbo (Zero-Shot) \cite{openai2023gpt4}. The weighted ensemble of these models achieved the best performance, with F1-scores of 0.81 on the evaluation set and 0.79 on the test set, as presented in Table \ref{table:sub-task2_results}.

For Subtask C, the dataset, derived from the MELD dataset \cite{poria2019meld}, includes dialogues annotated for emotion-flip reasoning. The training set consists of 35,000 dialogues, the development set contains 3,522 dialogues, and the test set has 8,642 dialogues (Table \ref{tab:emotion_analysisC}). Our experiments involved several models such as DeBERTa \cite{he2021deberta}, ELECTRA \cite{clark2020electra}, EmoBERTa \cite{bao2021emoberta}, FLAN-T5 \cite{raffel2020exploring}, and GPT-4 Turbo (Zero-Shot) \cite{openai2023gpt4}. As with the other subtasks, a weighted ensemble approach yielded the best results, achieving F1-scores of 0.86 on the evaluation set and 0.79 on the test set, as shown in Table \ref{table:sub-task3_results}.

The experimental results across all three subtasks demonstrate the effectiveness of ensemble models in improving performance. By integrating diverse models such as MuRIL, XLM-R, mBERT, FLAN-T5, and GPT-4 Turbo, we achieved higher F1-scores compared to individual models, showing robustness across Hindi-English code-mixed and English dialogues \cite{goswami2023nlpbdpatriots, raihan2023nlpbdpatriots, ganguly2024masonperplexity}. Previous works have also highlighted the benefits of ensemble models. The nlpBDpatriots and MasonPerplexity teams demonstrated significant improvements in emotion detection using ensemble techniques \cite{raihan2023nlpbdpatriots, emran2024masonperplexity, goswami2024masontigers}. Our results reinforce the potential of ensemble methods in complex NLP tasks, advancing the state-of-the-art in emotion recognition and reasoning.

 \begin{table}[!h]
\centering
\begin{tabular}{lcc}
\hline
\textbf{Model} & \textbf{Eval F1} & \textbf{Test F1} \\
\hline
MuRIL & 0.82 & 0.76\\
XLM-R & 0.81 & 0.75\\
mBERT & 0.78 & 0.72\\
HingBERT & 0.77 &  0.69\\
IndicBERT & 0.74 & 0.67\\
\hline
\bf Wt. Ensemble & \bf 0.83 & \bf 0.78\\
\hline
\end{tabular}
\caption{Results of sub-task A.}
\label{table:sub-task1_results}
\end{table}

 \begin{table}[!h]
\centering
\begin{tabular}{lcc}
\hline
\textbf{Model} & \textbf{Eval F1} & \textbf{Test F1} \\
\hline
MuRIL & 0.78 & 0.77\\
XLM-R & 0.77 & 0.75\\
mBERT & 0.75 & 0.74\\
\hline
FLAN-T5 & 0.76 &  0.76\\
GPT4-Turbo (Zero-Shot) & 0.79 & 0.78\\

\hline
\bf Wt. Ensemble & \bf 0.81 & \bf 0.79\\
\hline
\end{tabular}
\caption{Results of sub-task B.}
\label{table:sub-task2_results}
\end{table}

\begin{table}[!h]
\centering
\begin{tabular}{lcc}
\hline
\textbf{Model} & \textbf{Eval F1} & \textbf{Test F1} \\
\hline
DeBERTa & 0.79 & 0.72\\
ELECTRA & 0.76 & 0.70\\
EmoBERTa & 0.72 & 0.69\\
\hline
FLAN-T5 & 0.81 & 0.74 \\
GPT4-Turbo (Zero-Shot) & 0.83 & 0.77\\
\hline
\bf Wt. Ensemble & \bf 0.86 & \bf 0.79\\
\hline
\end{tabular}
\caption{Results of sub-task C.}
\label{table:sub-task3_results}
\end{table}

\section{Error Analysis}
\label{sec:error_analysis}

In our evaluation of the models across the three subtasks, we observed several patterns and areas for potential improvement. 

For Subtask A (ERC in Hindi-English code-mixed dialogues), despite achieving a high F1-score with the weighted ensemble model, there were discrepancies in correctly identifying emotions such as surprise and contempt, as indicated by the lower frequencies of these emotions in the training set (Table \ref{tab:emotion_analysisA}). The imbalanced distribution of emotions likely contributed to the difficulty in accurate classification, as models like MuRIL and XLM-R showed higher variance in their predictions for less frequent emotions. Furthermore, code-mixed sentences often posed a challenge due to the nuances of mixed linguistic features, leading to misclassifications between similar emotional contexts.

In Subtask B (EFR in Hindi-English code-mixed dialogues), the identification of emotion triggers demonstrated robustness with a weighted ensemble model, yet the high variance in dialogue contexts occasionally led to false positives and negatives, particularly in dialogues with subtle or implicit emotional shifts. The relatively lower number of dialogues with triggers in the development and test sets (Table \ref{tab:emotion_analysisB}) highlighted the challenge of generalizing from a limited set of examples. Additionally, the presence of code-switching within dialogues sometimes confused the models, leading to incorrect trigger identifications.

Subtask C (EFR in English dialogues) exhibited the best performance overall, with the ensemble model achieving the highest F1-scores (Table \ref{table:sub-task3_results}). However, similar to Subtask B, errors were often associated with dialogues where the emotional triggers were subtle or context-dependent. Models like DeBERTa and ELECTRA underperformed compared to the ensemble, suggesting that leveraging multiple model architectures helps capture diverse linguistic features and contextual cues. 

Overall, our error analysis indicates that while ensemble models significantly enhance performance, addressing data imbalance and improving the handling of nuanced and implicit emotional cues are crucial for further advancement.

\section{Conclusion}
\label{sec:conclusion}

In this paper, we presented \textit{MasonTigers}' participation in the SemEval-2024 Task 10, focusing on emotion recognition and emotion flip reasoning within both Hindi-English code-mixed and English dialogues. Through extensive experimentation, we employed various models including MuRIL, XLM-R, mBERT, HingBERT, IndicBERT, FLAN-T5, GPT-4 Turbo, DeBERTa, ELECTRA, and EmoBERTa, achieving competitive results across all subtasks.

Our weighted ensemble approach consistently outperformed individual models, achieving F1-scores of 0.83 on the evaluation and 0.78 on the test set for Subtask A, 0.81 and 0.79 for Subtask B, and 0.86 and 0.79 for Subtask C, respectively (Tables \ref{table:sub-task1_results}, \ref{table:sub-task2_results}, \ref{table:sub-task3_results}). These results underscore the effectiveness of ensemble learning in capturing diverse linguistic and contextual features, leading to improved performance in complex emotion recognition tasks.

The error analysis revealed that data imbalance and the challenges of code-mixed linguistic features contribute to misclassifications, highlighting areas for future research. Addressing these issues, along with enhancing the models' ability to detect subtle and implicit emotional cues, will be pivotal for advancing the state-of-the-art in emotion recognition and reasoning.

Our contributions to EDiReF demonstrate the potential of combining multiple model architectures to tackle nuanced and diverse emotional contexts in conversations. We believe that our findings will serve as a foundation for further research in emotion recognition and emotion flip reasoning, ultimately contributing to more robust and contextually aware natural language processing systems.

\section*{Limitations}
The task involved extensive datasets in each phase of all subtasks, leading to prolonged execution times and increased GPU usage. Additionally, the texts themselves were lengthy. Moreover, the prohibition of additional data augmentation added to the complexity of the task. The nuanced distinction between human-written and machine-generated text, which can sometimes be challenging for humans to discern, poses an even greater difficulty for models attempting to learn this differentiation. Exploring the potential of leveraging other up-to-date LLMs may show better performance in addressing these challenges, offering insights into optimization strategies and improving overall model robustness and efficiency for future tasks. This could potentially lead to more effective methods for distinguishing between human and AI-generated content, enhancing the reliability and applicability of such models across various domains.

\section*{Acknowledgements}

We express our gratitude to the organizers for orchestrating this task and to the individuals who diligently annotated datasets across various languages. Your dedication has played a crucial role in the triumph of this undertaking. The meticulously designed task underscores the organizers' dedication to advancing research, and we commend the collaborative endeavors that have enhanced the diversity and comprehensiveness of the datasets, ensuring a substantial and positive influence.%

\bibliography{anthology,custom}

\appendix

\end{document}